\documentclass[10pt,journal,compsoc]{IEEEtran}
\ifCLASSOPTIONcompsoc
  \usepackage[nocompress]{cite}
\else
  \usepackage{cite}
\fi
\ifCLASSINFOpdf
   \usepackage[pdftex]{graphicx}
\else
\fi
\usepackage{amsmath}
\usepackage{amssymb}
\usepackage{xcolor,tabulary,xfrac,enumitem}
\usepackage[ruled,noresetcount]{algorithm2e}
\usepackage[pagebackref=true,breaklinks=true,letterpaper=true,colorlinks,citecolor=citecolor,bookmarks=false]{hyperref}
\definecolor{citecolor}{RGB}{34,139,34}

\usepackage{array}

\ifCLASSOPTIONcompsoc
 \usepackage[caption=false,font=footnotesize,labelfont=sf,textfont=sf]{subfig}
\else
 \usepackage[caption=false,font=footnotesize]{subfig}
\fi
\usepackage{url}

\usepackage{booktabs}
\usepackage{multirow}
\usepackage{threeparttable}
\usepackage{color, colortbl}
\usepackage{changes}

\makeatletter
\newcommand{\thickhline}{%
    \noalign {\ifnum 0=`}\fi \hrule height 0.5pt
    \futurelet \reserved@a \@xhline
}
\makeatother
\definecolor{LightGray}{gray}{0.9}

\newcommand{\ie}{\textit{i.e.}}
\newcommand{\eg}{\textit{e.g.}}

\newcommand{\vs}{\textit{vs.}}
\newcommand{\pub}[1]{\color{gray}{\tiny{[{#1}]}}}

\begin{document}
%
\title{Collaborative Video Object Segmentation by Multi-Scale Foreground-Background Integration}
%
%
%
%

\newcommand{\zongxin}[1]{{#1}}

\author{Zongxin~Yang,
        Yunchao~Wei,
        and~Yi~Yang
\thanks{
Z. Yang, Y. Wei and Y. Yang are with the ReLER Lab, Centre for Artificial Intelligence, University of Technology Sydney, Ultimo, NSW 2007, Australia.
E-mail: zongxin.yang@student.uts.edu.au, \{yunchao.wei, yi.yang\}@uts.edu.au
}
}

\IEEEtitleabstractindextext{%
\begin{abstract}
This paper investigates the principles of embedding learning to tackle the challenging semi-supervised video object segmentation. Unlike previous practices that focus on exploring the embedding learning of foreground object (s), we consider background should be equally treated. Thus, we propose a Collaborative video object segmentation by Foreground-Background Integration (CFBI) approach. CFBI separates the feature embedding into the foreground object region and its corresponding background region, implicitly promoting them to be more contrastive and improving the segmentation results accordingly. Moreover, CFBI performs both pixel-level matching processes and instance-level attention mechanisms between the reference and the predicted sequence, making CFBI robust to various object scales. Based on CFBI, we introduce a multi-scale matching structure and propose an Atrous Matching strategy, resulting in a more robust and efficient framework, CFBI+. We conduct extensive experiments on two popular benchmarks, \ie, DAVIS, and YouTube-VOS. Without applying any simulated data for pre-training, our CFBI+ achieves the performance ($\mathcal{J}$\&$\mathcal{F}$) of 82.9\% and 82.8\%, outperforming all the other state-of-the-art methods.
Code: \url{https://github.com/z-x-yang/CFBI}.
\end{abstract}

}

\maketitle
\IEEEdisplaynontitleabstractindextext
%
\IEEEpeerreviewmaketitle

\IEEEraisesectionheading{\section{Introduction}\label{sec:introduction}}

Video Object Segmentation (VOS) is a fundamental task in computer vision with many potential applications, including augmented reality~\cite{ngan2011video} and self-driving cars~\cite{zhang2016instance}. In this paper, we focus on semi-supervised VOS, which targets on segmenting a particular object across the entire video sequence based on the object mask given at the first frame. 
The development of semi-supervised VOS can benefit many related tasks, such as video instance segmentation~\cite{vis,Feng_2019_ICCV} and interactive video object segmentation~\cite{oh2019fast,miao2020memory,liangmemory}.

Early VOS works(~\cite{osvos,onavos,premvos}) rely on fine-tuning with the first frame in evaluation, which heavily slows down the inference speed. Recent works (\eg,~\cite{osmn,feelvos,spacetime}) aim to avoid fine-tuning and achieve better run-time. 
In these works, STMVOS~\cite{spacetime} introduces memory networks to learn to read sequence information and outperforms all the fine-tuning based methods. However, STMVOS relies on simulating extensive frame sequences using large image datasets~\cite{voc,coco,cheng2014global,shi2015hierarchical,semantic} for training. The simulated data significantly boosts the performance of STMVOS but makes the training procedure elaborate. Without simulated data, FEELVOS~\cite{feelvos} adopts a semantic pixel-wise embedding together with a global (between the first and current frames) and a local (between the previous and current frames) matching mechanism to guide the prediction. The matching mechanism is simple and fast, but the performance is not comparable with STMVOS.

\begin{figure}[t!]
\centering 

\includegraphics[width=0.98\linewidth]{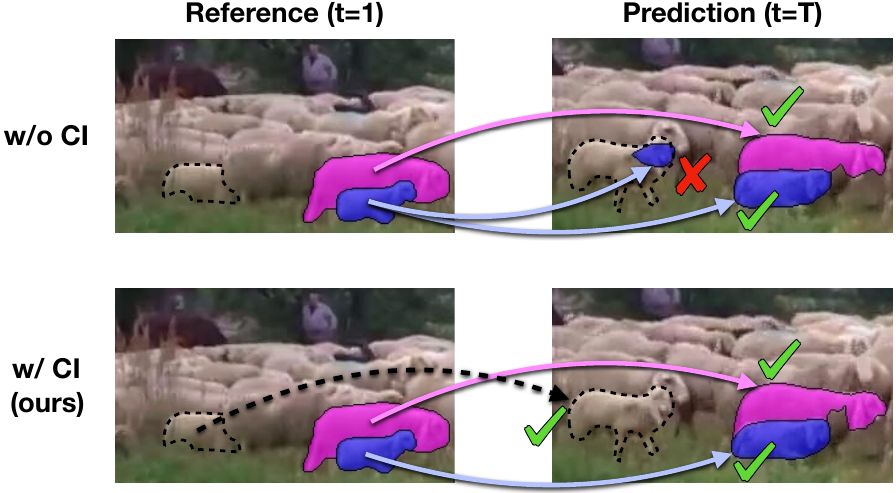}

\caption{CI: collaborative integration. There are two foreground sheep (pink and blue) in the sequence. In the top line, the contempt of background matching leads to a confusion of sheep's prediction. In the bottom line, we relieve the confusion problem by introducing background matching (dot-line arrow).}\label{fig:cl}

\end{figure}

Even though the efforts mentioned above have made significant progress, current state-of-the-art works pay little attention to the feature embedding of background region in videos and only focus on exploring robust matching strategies for the foreground object (s). Intuitively, it is easy to extract the foreground region from a video when precisely removing all the background. Moreover, modern video scenes commonly focus on many similar objects, such as the cars in car racing, the people in a conference, and the animals on a farm. For these cases, the contempt of integrating foreground and background embeddings traps VOS in an unexpected background confusion problem. As shown in Fig.~\ref{fig:cl}, if we focus on only the foreground matching like FEELVOS, a similar and same kind of object (sheep here) in the background is easy to confuse the prediction of the foreground object. Such an observation motivates us that the background should be equally treated compared with the foreground so that better feature embedding can be learned to relieve the background confusion and promote the accuracy of VOS.

\zongxin{We propose a novel framework for Collaborative video object segmentation by Foreground-Background Integration (CFBI) based on the above motivation. 
Unlike the above methods, we not only extract the embedding and do matching for the foreground target in the reference frame, but also for the background region to relieve the background confusion.
In particular, our framework extracts two types of embedding, pixel-level and instance-level, for each video frame to cover different scales of features. Like FEELVOS, we employ pixel-level embedding to match all the objects' details with the same global \& local mechanism. However, the pixel-level matching is not sufficient and robust to match those objects with larger scales and may bring unexpected noises due to the pixel-wise diversity. Thus we introduce instance-level embedding to help the segmentation of large-scale objects by using attention mechanisms.
For the training process, we propose a balanced random-crop scheme to avoid biasing learned attributes to background attributes.
These proposed strategies can effectively improve the quality of the learned collaborative embeddings for conducting VOS while keeping the network simple yet effective simultaneously.}
\zongxin{Based on CFBI, we further introduce an efficient multi-scale matching structure, resulting in a more robust framework, CFBI+. Within CFBI+, we propose an Atrous Matching (AM) strategy, which can significantly save computation and memory usage of matching processes. The use of AM makes CFBI+ not only more robust but also more efficient than CFBI.}

{We perform extensive experiments on DAVIS~\cite{davis2016,davis2017}, and YouTube-VOS~\cite{youtubevos} to validate the effectiveness of the proposed CFBI and CFBI+. Without any bells and whistles (such as the use of simulated data, fine-tuning, or post-processing), CFBI+ outperforms all other state-of-the-art methods on the validation splits of DAVIS 2017 (ours, $\mathcal{J}\&\mathcal{F}$ $\mathbf{82.9\%}$) and YouTube-VOS ($\mathbf{82.8\%}$). 
Meanwhile, our multi-object inference speed is faster than previous state-of-the-art methods.
We have made the code publicly available, and we hope our simple yet effective CFBI and CFBI+ will serve as two solid baselines and help ease the future research related to VOS.}

{This paper is an extension of our previous conference version~\cite{cfbi}. 
The current work adds to the initial version in some significant aspects. 
First, we propose a plug-and-play Atrous Matching (AM) algorithm, which can significantly save computation and memory usage of matching processes.
Second, based on the proposed AM, we design a multi-scale matching framework, resulting in a more strong and efficient VOS framework, CFBI+. Third,  we incorporate considerable new experimental results, including ablation study, model setting, and visualization analysis.
}

\section{Related Work}\label{sec:related_work}

\noindent\textbf{Semi-supervised Video Object Segmentation.}
Many previous methods for semi-supervised VOS rely on fine-tuning at test time. Among them, OSVOS~\cite{osvos} and MoNet~\cite{xiao2018monet} fine-tune the network on the first-frame ground-truth at test time. OnAVOS~\cite{onavos} extends the first-frame fine-tuning by an online adaptation mechanism, \ie, online fine-tuning. MaskTrack~\cite{masktrack} uses optical flow to propagate the segmentation mask from one frame to the next. PReMVOS~\cite{premvos} combines four different neural networks (including an optical flow network~\cite{flownet}) using extensive fine-tuning and a merging algorithm. Despite achieving promising results, all these methods are seriously slowed down by fine-tuning during inference.

Some other recent works (\eg,~\cite{osmn,favos}) aim to avoid fine-tuning and achieve a better run-time. OSMN~\cite{osmn} employs two networks to extract the instance-level information and make segmentation predictions, respectively.  PML~\cite{pml} learns a pixel-wise embedding with the nearest neighbor classifier. Similar to PML, VideoMatch~\cite{videomatch} uses a soft matching layer that maps the pixels of the current frame to the first frame in a learned embedding space. Following PML and VideoMatch, FEELVOS~\cite{feelvos} extends the pixel-level matching mechanism by additionally matching between the current frame and the previous frame. Compared to fine-tuning methods, FEELVOS achieves a much higher speed, but there is still an accuracy gap. Like FEELVOS, RGMP~\cite{rgmp} and STMVOS~\cite{spacetime} does not require any fine-tuning. STMVOS, which leverages a memory network to store and read the information from past frames, outperforms all the previous methods. {However, STMVOS and its following works (EGMN~\cite{EGMN} and KMNVOS~\cite{KMN}) rely on an elaborate pre-training procedure using extensive simulated data generated from multiple datasets with pixel-level annotations. LWLVOS~\cite{LWLVOS} proposes to use an online few-shot learner during both training and testing stages. Without simulated data, LWLVOS is comparable with KMNVOS on YouTube-VOS, but generalizes worse than the above methods using simulated data on DAVIS.}

{In previous practices, learning foreground feature embedding has been well explored. OSMN proposed to conduct an instance-level matching, but such a matching scheme fails to consider the feature diversity among the details of the target's appearance and results in coarse predictions. PML and FEELVOS alternatively adopt the pixel-level matching by matching each pixel of the target, which effectively takes the feature diversity into account and achieves promising performance. Nevertheless, performing pixel-level matching may bring unexpected noises in the case of some pixels from the background are with a similar appearance to the ones from the foreground (Fig.~\ref{fig:cl}).}

\begin{figure*}[t!]
    \centering
    \includegraphics[width=0.9\linewidth]{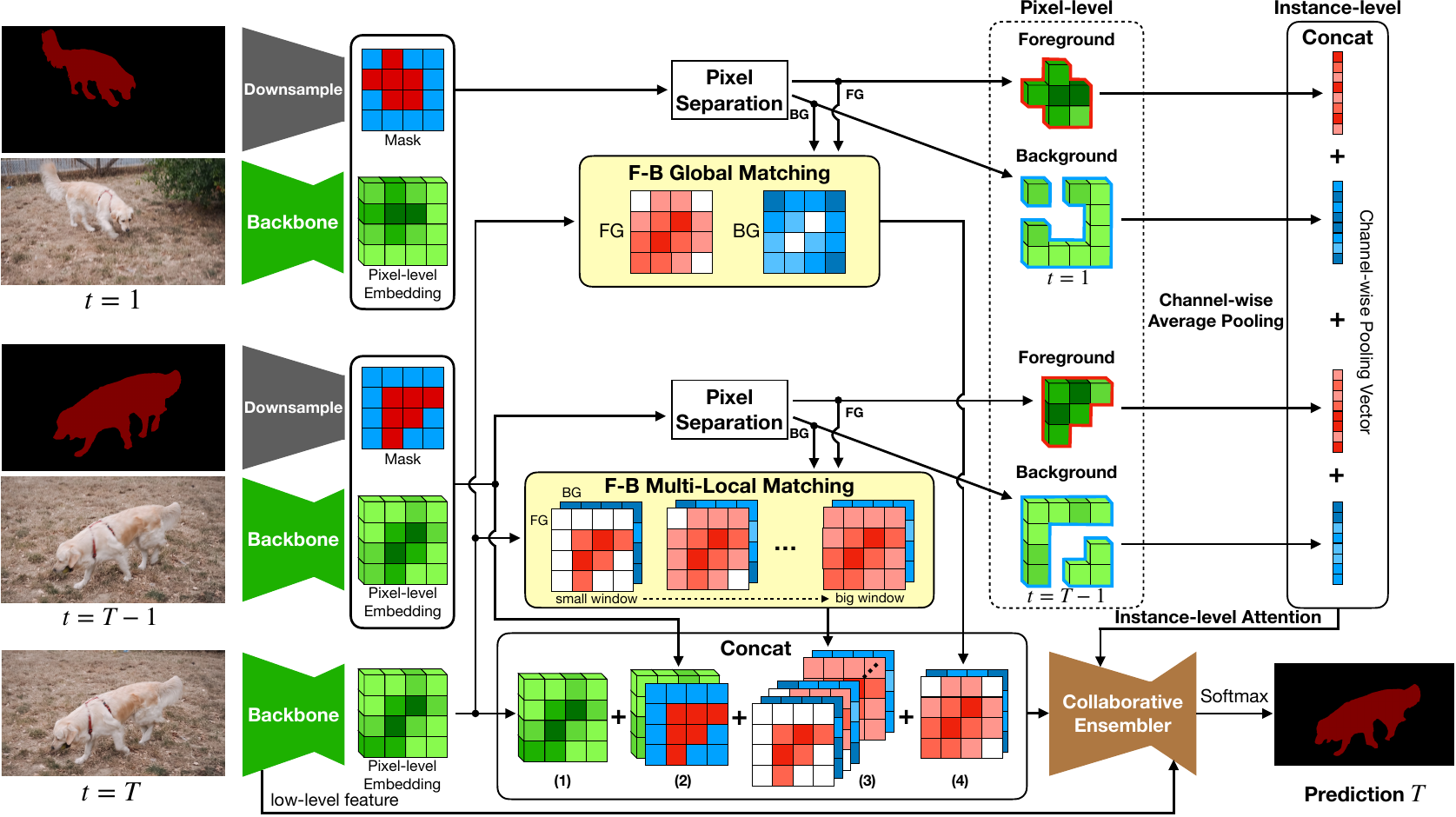}
    \caption{An \textbf{overview} of CFBI. F-G denotes Foreground-Background. We use \textcolor{red}{red} and \textcolor{blue}{blue} to indicate foreground and background separately. The deeper the red or blue color, the higher the confidence. Given the first frame ($t=1$), previous frame ($t=T-1$), and current frame ($t=T$), we firstly extract their pixel-wise embedding by using a backbone network. Second, we separate the first and previous frame embeddings into the foreground and background pixels based on their masks. After that, we use F-G pixel-level matching and instance-level attention to guide our collaborative ensembler network to generate a prediction.}
    \label{fig:overview}
\end{figure*}

{Thus, we propose a collaborative integration method by additionally learning background embedding. 
Furthermore, our CFBI utilizes both the pixel-level and instance-level embeddings to guide prediction.}

\noindent\textbf{Attention Mechanisms.}
Recent works introduce the attention mechanism into convolutional networks (\eg, ~\cite{attention_conv1,attention_conv2}). 
Following them, SE-Nets~\cite{senet} introduced a lightweight gating mechanism that focuses on enhancing the representational power of the convolutional network by modeling channel attention. Inspired by SE-Nets, CFBI uses an instance-level average pooling method to embed collaborative instance information from pixel-level embeddings. After that, we conduct a channel-wise attention mechanism to help guide prediction. Compared to OSMN, which employs an additional convolutional network to extract instance-level embedding, our instance-level attention method is more efficient and lightweight.

\section{Methodology}\label{sec:methodology}

\noindent\textbf{The Overview of CFBI.} 
To overcome or relieve the problems raised by previous methods and promote the foreground objects from the background, we present Collaborative video object segmentation by Foreground-Background Integration (CFBI), as shown in Figure~\ref{fig:overview}. We use \textcolor{red}{red} and \textcolor{blue}{blue} to indicate foreground and background separately. First, beyond learning feature embedding from foreground pixels, our CFBI also considers embedding learning from background pixels for collaboration. Such a learning scheme will encourage the feature embedding from the target object and its corresponding background to be contrastive, promoting the segmentation results accordingly. Second, we further conduct the embedding matching from both pixel-level and instance-level with the collaboration of pixels from the foreground and background. For the pixel-level matching, 
we improve the robustness of the local matching under various object moving rates. For the instance-level matching, we design an instance-level attention mechanism to augment the pixel-level matching efficiently. Moreover, to implicitly aggregate the learned foreground \& background and pixel-level \& instance-level information, we employ a collaborative ensembler to construct large receptive fields and make precise predictions.

\subsection{Collaborative Pixel-level Matching}

For the pixel-level matching, we adopt a global and local matching mechanism similar to FEELVOS for introducing the guided information from the first and previous frames, respectively. Unlike previous methods~\cite{pml,feelvos}, we additionally incorporate background information and apply multiple windows in the local matching, which is shown in the middle of Fig.~\ref{fig:overview}. 

For incorporating background information, we firstly redesign the pixel distance of~\cite{feelvos} to further distinguish the foreground and background.
Let $B_t$ and $F_t$ denote the pixel sets of background and all the foreground objects of frame $t$, respectively. We define a new distance between pixel $p$ of the current frame $T$ and pixel $q$ of frame $t$ in terms of their corresponding embedding, $e_p$ and $e_q$, by
\begin{equation} \label{equ:distance}
    D(p,q)=
        \begin{cases}
            1-\frac{2}{1+exp(||e_p-e_q||^2+b_B)} & \text{if } q \in B_t\\
            1-\frac{2}{1+exp(||e_p-e_q||^2+b_F)} & \text{if } q \in F_t
        \end{cases},
\end{equation}
where $b_B$ and $b_F$ are trainable background bias and foreground bias. We introduce these two biases to make our model be able further to learn the difference between foreground distance and background distance.

\begin{figure*}[t!]
\center

\subfloat[Slow moving rate]{
\label{fig:slow}
\includegraphics[width=0.4\linewidth]{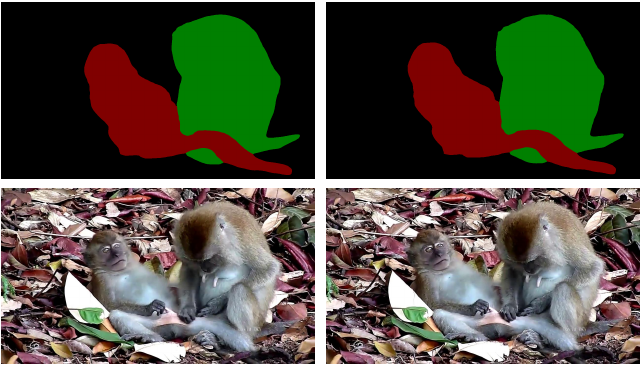}
}
\subfloat[Fast moving rate]{
\label{fig:fast}
\includegraphics[width=0.4\linewidth]{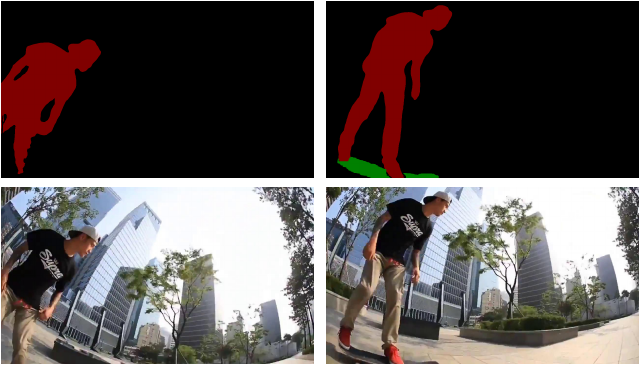}
}

\caption{The moving rate of objects across two adjacent frames is largely variable for different sequences. Examples are from YouTube-VOS~\cite{youtubevos}.}\label{fig:offset}
\end{figure*}

\noindent\textbf{Foreground-Background Global Matching.} Let $\mathcal{P}_t$ denote the set of all pixels (with a stride of 4) at time $t$ and $\mathcal{P}_{t,o}\subseteq \mathcal{P}_{t}$ is the set of pixels at time $t$ which belongs to the foreground object $o$. The global foreground matching between one pixel $p$ of the current frame $T$ and the pixels of the first reference frame (\ie, $t=1$) is,
\begin{equation} \label{equ:global_f}
    G_{o}(p)=\min_{q\in\mathcal{P}_{1,o}} D(p,q).
\end{equation}
Similarly, let $\mathcal{\overline{P}}_{t,o} =\mathcal{P}_t \backslash \mathcal{P}_{t,o}$ denote the set of relative background pixels of object $o$ at time $t$, and the global background matching is,
\begin{equation} \label{equ:global_b}
    \overline{G}_{o}(p)=\min_{q\in\mathcal{\overline{P}}_{1,o}} D(p,q).
\end{equation}

\noindent\textbf{Foreground-Background Multi-Local Matching.}
In FEELVOS, the local matching is limited in only one fixed extent of neighboring pixels, but the offset of objects across two adjacent frames in VOS is variable, as shown in Fig.~\ref{fig:offset}. Thus, we propose to apply the local matching mechanism on different scales and let the network learn how to select an appropriate local scale, which makes our framework more robust to various moving rates of objects. Notably, we use the intermediate results of the local matching with the largest window to calculate on other windows. Thus, the increase of computational resources of our multi-local matching is negligible.

Formally, let $K=\{k_1,k_2,...,k_n\}$ denote all the neighborhood sizes and $H(p,k)$ denote the neighborhood set of pixels that are at most $k$ pixels away from $p$ in both $x$ and $y$ directions, our foreground multi-local matching between the current frame $T$ and its previous frame $T-1$ is
\begin{equation} \label{equ:multi_local_f}
    ML_{o}(p,K)=\{L_{o}(p,k_1),L_{o}(p,k_2),...,L_{o}(p,k_n)\},
\end{equation}
where
\begin{equation} \label{equ:local_f}
    L_{o}(p,k)=
        \begin{cases}
            \min_{q\in\mathcal{P}^{p,k}_{T-1,o}} D(p,q) & \text{if }\mathcal{P}^{p,k}_{T-1,o}\neq\emptyset \\
            1 & \text{otherwise}
        \end{cases}.
\end{equation}
Here, $\mathcal{P}^{p,k}_{T-1,o}:=\mathcal{P}_{T-1,o}\cap H(p,k)$ denotes the pixels in the local window (or neighborhood). And our background multi-local matching is
\begin{equation} \label{equ:multi_local_b}
    \overline{ML}_{o}(p,K)=\{\overline{L}_{o}(p,k_1),\overline{L}_{o}(p,k_2),...,\overline{L}_{o}(p,k_n)\},
\end{equation}
where
\begin{equation} \label{equ:local_b}
    \overline{L}_{o}(p,k)=
        \begin{cases}
            \min_{q\in\mathcal{\overline{P}}_{T-1,o}^{p,k}} D(p,q) & \text{if }\mathcal{\overline{P}}_{T-1,o}^{p,k}\neq\emptyset \\
            1 & \text{otherwise}
        \end{cases}.
\end{equation}
Here similarly, $\mathcal{\overline{P}}^{p,k}_{T-1,o}:=\mathcal{\overline{P}}_{T-1,o}\cap H(p,k)$.

In addition to the global and multi-local matching maps, we concatenate the pixel-level embedding feature and mask of the previous frame with the current frame feature. FEELVOS demonstrates the effectiveness of concatenating the previous mask. Following this, we empirically find that introducing the previous embedding can further improve the performance ($\mathcal{J}$\&$\mathcal{F}$) by about $0.5\%$.

In summary, the output of our collaborative pixel-level matching is a concatenation of (1) the pixel-level embedding of the current frame, (2) the pixel-level embedding and mask of the previous frame, (3) the multi-local matching map and (4) the global matching map, as shown in the bottom box of Fig.~\ref{fig:overview}.

\begin{figure}[t!]
\center

\includegraphics[width=0.47\linewidth]{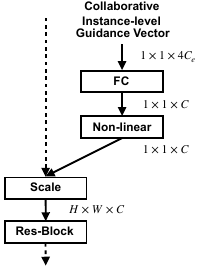}

\caption{The trainable part of the instance-level attention. $C_e$ denotes the channel dimension of pixel-wise embedding. $H$, $W$, $C$ denote the height, width, channel dimension of CE features.}
\label{fig:instance}

\end{figure}

\subsection{Collaborative Instance-level Attention}

As shown in the right of Fig~\ref{fig:overview}, we further design a Collaborative instance-level attention mechanism to guide the segmentation for large-scale objects. 

After getting the pixel-level embeddings of the first and previous frames, we separate them into foreground and background pixels (\ie, $\mathcal{P}_{1,o}$, $\mathcal{\overline{P}}_{1,o}$, $\mathcal{P}_{T-1,o}$, and $\mathcal{\overline{P}}_{T-1,o}$) according to their masks. Then, we apply channel-wise average pooling on each group of pixels to generate a total of four instance-level embedding vectors and concatenate these vectors into one collaborative instance-level guidance vector. Thus, the guidance vector contains the information from both the first and previous frames, and both the foreground and background regions.

\begin{figure*}[ht!]
\center
\includegraphics[width=0.8\linewidth]{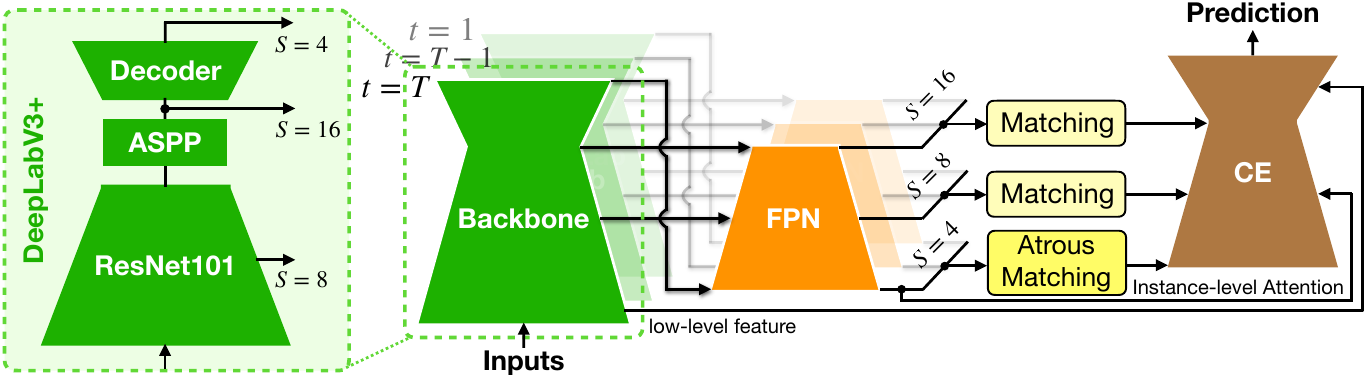}
\caption{An overview of CFBI+. $S$: the stride of feature maps. Firstly, CFBI+ extracts three features with different scales ($S=4,8,16$) from backbone, ResNet101-DeepLabV3+~\cite{deeplabv3p}. And then, we use the Feature Pyramid Network (FPN)~\cite{fpn} to fuse the information from small scales to large scales and reduce the channel dimensions of three features. After this, we do all the matching processes of CFBI on each scale. The output of each scale will be sent to the consistent stage of Collaborative Ensembler (CE).}\label{fig:ms_matching}
\end{figure*}

In order to efficiently utilize the instance-level information, we employ an attention mechanism to adjust our Collaborative Ensembler (CE). We show a detailed illustration in Fig.~\ref{fig:instance}. Inspired by SE-Nets~\cite{senet}, we leverage a fully-connected (FC) layer (we found this setting is better than using two FC layers as adopted by SE-Net) and a non-linear activation function to construct a gate for the input of each Res-Block in the CE. The gate will adjust the scale of the input feature channel-wisely.

By introducing collaborative instance-level attention, we can leverage a full scale of foreground-background information to guide the prediction further. The information with a large (instance-level) receptive field is useful to relieve local ambiguities~\cite{torralba2003contextual}, which is inevitable with a small (pixel-wise) receptive field.

\begin{figure*}[ht!]
\center

\subfloat[$l=1$ (original matching)]{
\includegraphics[width=0.3\linewidth]{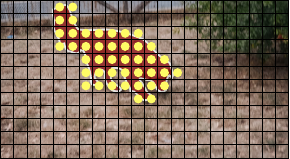}
}
\subfloat[$l=2$]{
\includegraphics[width=0.3\linewidth]{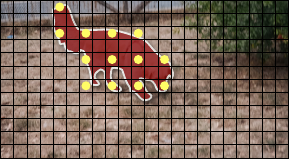}
}
\subfloat[$l=3$]{
\includegraphics[width=0.3\linewidth]{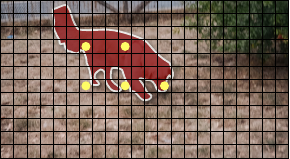}
}

\caption{An illustration of $l$-atrous object pixel set. Atrous matching improves computational efficiency by periodically filtering out referred object pixels without loss of resolution. (a) Yellow points indicate the referred object pixel set used in the original matching process. All the object (red dog) pixels are sampled. (b) (c) $2$-atrous and $3$-atrous object pixel set. Referred pixels are sampled out with a period of $2$ or $3$ pixels vertically and horizontally.}\label{fig:atrous_matching}
\end{figure*}

\subsection{Collaborative Ensembler (CE)}

In the lower right of Fig.~\ref{fig:overview}, we design a collaborative ensembler for making large receptive fields to aggregate pixel-level and instance-level information and implicitly learn the collaborative relationship between foreground and background. 

Inspired by ResNets~\cite{resnet} and Deeplabs~\cite{deeplab,deeplabv3p}, which both have shown significant representational power in image segmentation tasks, our CE uses a downsample-upsample structure, which contains three stages of Res-Blocks~\cite{resnet} and an Atrous Spatial Pyramid Pooling (ASPP)~\cite{deeplabv3p} module. The number of Res-Blocks in Stage 1, 2, and 3 are $2$, $3$, $3$ in order. Besides, we employ dilated convolutional layers to improve the receptive fields efficiently. The dilated rates of the $3\times3$ convolutional layer of Res-Blocks in one stage are separately $1$, $2$, $4$ ( or $1$, $2$ for Stage 1). At the beginning of Stage 2 and Stage 3, the feature maps will be downsampled by the first Res-Block with a stride of 2. After these three stages, we employ an ASPP and a Decoder~\cite{deeplabv3p} module to increase the receptive fields further, upsample the scale of feature and fine-tune the prediction collaborated with the low-level backbone features.

\subsection{CFBI+: Towards Efficient Multi-scale Matching}
High-quality matching maps are essential for CFBI to generate accurate predictions with sharp object boundaries, which is one of the critic factors for improving VOS's performance.
{However, it is costly in terms of both GPU memory and time to achieve a delicate and high-resolution matching map (\eg, with a stride of 4).}
Intuitively, there are two ways to accelerate matching processes. (1) Performing matching over low-resolution feature maps. Even the process will be much more light-weight, it is easy to miss many object details. (2) Reducing the channel dimensions while keeping using a high-resolution feature map. Even the amount of computation will decrease linearly with the channel dimensions, the accuracy of matching will also decrease.

Some recent works (\eg,~\cite{fpn,pspnet}) prove that multi-scale strategies can efficiently improve convolutional networks' performance. We consider that such strategies also benefit matching processes. Thus, we introduce an efficient multi-scale matching structure into CFBI, resulting in a more robust framework, \ie, CFBI+. An overview of CFBI+ is shown in Fig.~\ref{fig:ms_matching}. Firstly, CFBI+ extracts three features with different scales ($S=4,8,16$) from backbone. {And then, we use the Feature Pyramid Network (FPN)~\cite{fpn} to further fuse the information from small scales to large scales.} After this, we do all the matching processes of CFBI on every scale. The output of each scale will be sent to each corresponding stage of CE.

{Concretely, to harness the merits of the two accelerate matching processes as well as alleviate their disadvantages, we adopt an adaptive matching strategy for feature maps of different scales. CFBI+ progressively and linearly increases the channel dimensions from larger scales to smaller scales, which reduces the amount of calculation for matching on larger scales. Meanwhile, richer semantic information in smaller scales can successfully make up for the performance drop due to the reduction of channel dimension for larger scales. In this way, various coarse-to-fine information can help CFBI+ achieve better segmentation results.}

{Besides, only progressively and linearly increasing channel dimensions is not enough to make the multi-scale matching more efficient than single-scale, because the complexity of matching processes increases exponentially with the resolution of feature maps. For example, the calculation of $G_{T,o}(p)$ on $S=4$ scale is $256$ times of $S=16$. Thus, we additionally propose an Atrous Matching (AM) strategy to save computation and memory usage of matching processes further. Introducing AM helps CFBI+ to be more efficient than CFBI.}

\noindent\textbf{Atrous Matching (AM).} 
\textit{Algorithme \`atrous} (or "atrous algorithm" in the following text), an algorithm for wavelet decomposition~\cite{holschneider1990real}, played a key role in some recent convolutional networks~\cite{dilated_conv,deeplabv3p}. By adding a spatial interval during sampling, the atrous algorithm can reduce calculation while maintaining the same resolution. Intuitively, spatially close pixels always share similar semantic information. Hence, we argue that the atrous algorithm is also effective in matching processes. Removing part of similar pixels from referred pixels will not heavily drop performance but save much computation cost. 

Let $q_{x,y}$ be the pixel at position $(x,y)$ and let $l$ be an atrous factor, we generalize the foreground global matching (Eq. \ref{equ:global_f}) into an atrous form,
\begin{equation} \label{equ:atrous_global_f}
    G^{l}_{o}(p)=\min_{q\in\mathcal{P}^{l}_{1,o}} D(p,q),
\end{equation}
where 
\begin{equation}
    \mathcal{P}^{l}_{1,o}=\{q_{x,y}\in \mathcal{P}_{1,o},\forall x,y\in \{l, 2l, 3l, ...\}\}
\end{equation}
is a $l$-atrous object pixel set. We show an illustration in Fig.~\ref{fig:atrous_matching}.

Let $x_p$ and $y_p$ denote the position of pixel $p$, the atrous form of the foreground local matching (Eq.~\ref{equ:local_f}) is
\begin{equation} \label{equ:atrous_local_f}
    L^{l}_{o}(p,k)=
        \begin{cases}
            \min_{q\in\mathcal{P}^{l,p,k}_{T-1,o}} D(p,q) & \text{if }\mathcal{P}^{l,p,k}_{T-1,o}\neq\emptyset \\
            1 & \text{otherwise}
        \end{cases},
\end{equation}
where
\begin{equation}
    \mathcal{P}^{l,p,k}_{T-1,o}:=\mathcal{P}_{T-1,o}\cap H^{l}(p,k),
\end{equation}
and
\begin{equation}
\begin{aligned}
    H^{l}(p,k) = \{q_{x,y}\in H(p,k) \forall x\in \{x_p, x_p\pm l,  x_p\pm 2l, ...\}, \\ 
     y\in \{y_p, y_p\pm l, y_p\pm 2l, ...\}\}
\end{aligned}
\end{equation}
is a $l$-atrous neighborhood set.

In the same way, we can also generalize Eq.~\ref{equ:global_b}, Eq.~\ref{equ:local_b} Eq.~\ref{equ:multi_local_f}, and Eq.~\ref{equ:multi_local_b} into atrous forms, \ie, $\overline{G}^{l}_{o}(p)$, $\overline{L}^{l}_{o}(p,k)$, $ML^{l}_{o}(p,K)$, and $\overline{ML}^{l}_{o}(p,K)$.
Since the number of referred pixels is reduced $l^2$ times, AM's computational complexity is only $1/l^2$ of original matching.
Notably, AM is equivalent to original matching when $l$ is equal to $1$, \ie, $G^{l=1}_{o}(p)\equiv G_{o}(p)$ and $L^{l=1}_{o}(p,k)\equiv L_{o}(p,k)$.

{On the largest matching scale ($S=4$) of CFBI+, we apply $2$-atrous matching processes, which significantly improve the efficiency of CFBI+.
Notably, AM is a plug-and-play algorithm and can also improve the efficiency of CFBI during the testing stage.}

\section{Implementation Details}

\begin{figure}[t!]
\center

\subfloat[Normal]{
\label{fig:normal_crop}
\includegraphics[width=0.426\linewidth]{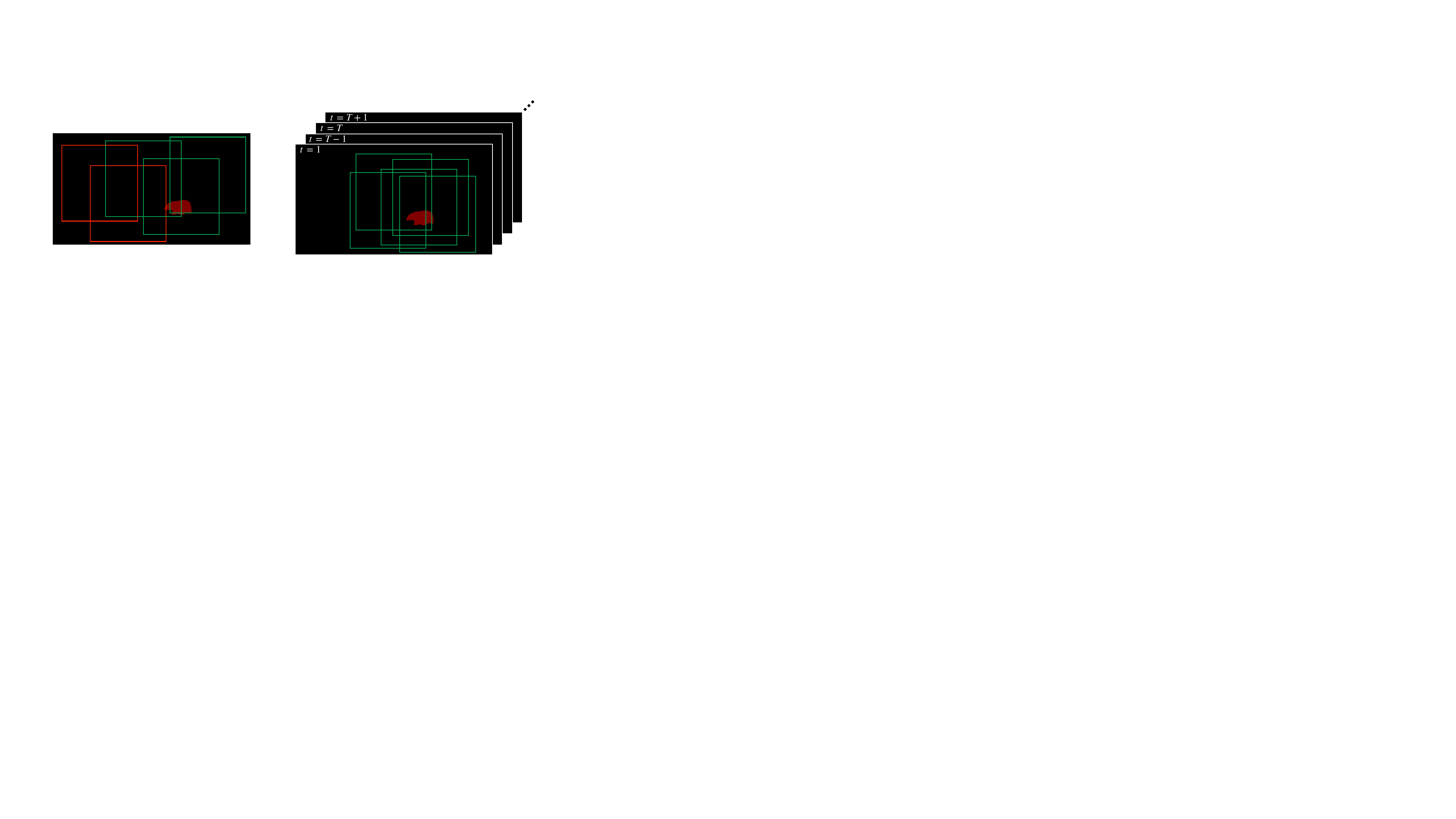}
}
\subfloat[Balanced]{
\label{fig:balanced_crop}
\includegraphics[width=0.52\linewidth]{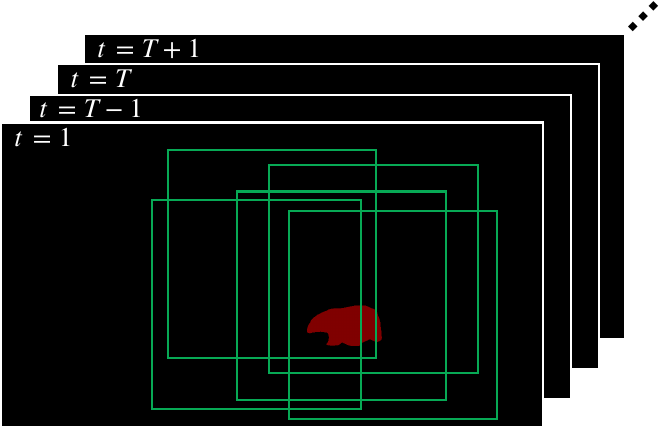}
}

\caption{When using normal random-crop, some red windows contain few or no foreground pixels. For reliving this problem, we propose balanced random-crop.}\label{fig:crop}

\end{figure}

\begin{figure*}[!t]
    \center
        \includegraphics[width=0.66\linewidth]{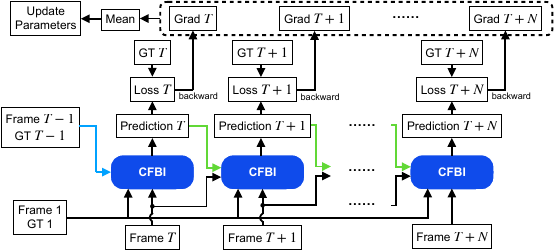}
        \caption{An illustration of the sequential training. In each step, the previous mask comes from the previous prediction (the green lines) except for the first step, whose previous mask comes from the ground-truth (GT) mask (the blue line).}
        \label{fig:sequence_training}
\end{figure*}

Following FEELVOS, we use the DeepLabv3+~\cite{deeplabv3p} architecture as the backbone for our network. However, our backbone is based on the dilated ResNet-101~\cite{deeplabv3p} instead of Xception-65~\cite{xception} for saving computational resources. We apply batch normalization (BN)~\cite{bn} in our backbone and pre-train it on ImageNet~\cite{deng2009imagenet} and COCO~\cite{coco}. To make the training process more effective and consistent with the inference stage, we additionally adopt two tricks, \ie, Balanced Random-Crop, and Sequential Training:
\begin{itemize}
    \item \textbf{Balanced Random-Crop.}
As shown in Fig.~\ref{fig:crop}, there is an apparent imbalance between the foreground and the background pixel number on VOS datasets. Such an issue usually makes the models easier to be biased to background attributes. 
In order to relieve this problem, we take a balanced random-crop scheme, which crops a sequence of frames (\ie, the first frame, the previous frame, and the current frame) by using a same cropped window and restricts the cropped region of the first frame to contain enough foreground information. The restriction method is simple yet effective. To be specific, the balanced random-crop will decide on whether the randomly cropped frame contains enough pixels from foreground objects or not. If not, the method will continually take the cropping operation until we obtain an expected one.
\item \textbf{Sequential Training.} In the training stage, FEELVOS predicts only one step in one iteration, and the guidance masks come from the ground-truth data. RGMP and STMVOS use previous guidance information (mask or feature memory) in training, which is more consistent with the inference stage and performs better. The previous guidance masks are always generated by the network in the previous inference steps in the evaluation stage. Following RGMP, we train the network using a sequence of consecutive frames in each SGD iteration. In each iteration, we randomly sample a batch of video sequences. For each video sequence, we randomly sample a frame as the reference frame and a continuous $N+1$ frames as the previous frame and current frame sequence (with $N$ frames). When predicting the first frame, we use the ground-truth of the previous frame as the previous mask. When predicting the following frames, we use the latest prediction to be the previous mask. We show an illustration in Fig.~\ref{fig:sequence_training}.
\end{itemize}

In CFBI, the backbone is followed by one depth-wise separable convolution for extracting pixel-wise embedding (channels=100) with a stride of 4. We further downsample the embedding feature to a half size for the multi-local matching using bi-linear interpolation for saving GPU memory.

In CFBI+, the backbone is followed by FPN~\cite{fpn} for extracting three pixel-wise embeddings (channels=32, 64, and 128) with strides of 4, 8, and 16, respectively. The window sizes are $\{4, 8, 12, 16, 20, 24\}$, $\{2, 4, 6, 8, 10, 12\}$, and $\{4, 6, 8, 10\}$ for three scales (stride$=4$, $8$, and $16$).

For the collaborative ensembler, we apply Group Normalization (GN)~\cite{gn} and Gated Channel Transformation (GCT)~\cite{gct} to improving training stability and performance when using a small batch size. We initialize $b_B$ and $b_F$ to $0$. In CFBI+, each matching scale has individual $b_B$ and $b_F$.

\begin{table}[t!]
	\centering
	\caption{The quantitative evaluation on YouTube-VOS~\cite{youtubevos}. F, S, and $^*$ denote online fine-tuning, using simulated data during training and performing model ensemble in evaluation, respectively. t/s: time per frame in seconds. $^{2\times}$: using double batch size and learning rate during training. $^{MS}$: using a multi-scale and flip strategy in evaluation. 
	}\label{tab:youtubevos}
	\setlength{\tabcolsep}{4pt}
	\begin{tabular}{lcccccccc}
	
\toprule[1.5pt]
            &  &  &  &  \multicolumn{2}{c}{Seen}  &    \multicolumn{2}{c}{Unseen} &  \\
\midrule[1pt]
 Methods & F & S  & Avg & $\mathcal{J}$ & $\mathcal{F}$ & $\mathcal{J}$ & $\mathcal{F}$ & t/s \\
\midrule[1pt]
\multicolumn{9}{c}{\textit{Validation 2018 Split}} \\
\midrule[1pt]
A-GAME~\pub{CVPR19}~\cite{agame} &   &   &  66.1  &  67.8  &  -  &  60.8  &  - & - \\
PReMVOS~\pub{ACCV18}~\cite{premvos} & \checkmark &   &  66.9  &  71.4  &  75.9  &  56.5  &  63.7 & 6 \\
BoLT~\pub{arXiv19}~\cite{boltvos} & \checkmark &   &  71.1  &  71.6  &  -  &  64.3  &  - & 1.36 \\
STMVOS~\pub{ICCV19}~\cite{spacetime} &  &   &  68.2  &  -  &  -  &  -  &  - & - \\
STMVOS~\pub{ICCV19}~\cite{spacetime} &  & \checkmark  &  79.4  &  79.7  &  84.2  &  72.8  &  80.9 & - \\
EGMN~\pub{ECCV20}~\cite{EGMN} &  & \checkmark  &  80.2  &  80.7  &  85.1  &  74.0  &  80.9 & - \\
KMNVOS~\pub{ECCV20}~\cite{KMN} &  & \checkmark  &  81.4  &  81.4  &  85.6  &  75.3  &  83.3 & - \\
LWLVOS~\pub{ECCV20}~\cite{LWLVOS} &  &  & 81.5  &  80.4  &  84.9  &  76.4  &  84.4 & - \\
\hline
CFBI &  &   &  81.4  &  81.1  & 85.8  & 75.3  & 83.4 & 0.29\\
CFBI$^{2\times}$ &  &   &  81.8  &  \textbf{81.9}  & 86.3  & 75.6  & 83.4 & 0.29\\
CFBI+  &  &   &  82.0  &  81.2  & 86.0  & 76.2  & 84.6 & \textbf{0.25} \\
CFBI+$^{2\times}$  &  &   &  \textbf{82.8}  &  81.8  & \textbf{86.6}  & \textbf{77.1}  & \textbf{85.6} & \textbf{0.25} \\
CFBI$^{MS}$ &  &   &  82.7  &  82.2  & 86.8  & 76.9  & 85.0 & 2.51 \\
CFBI+$^{MS}$ &  &   &  \textbf{83.3}  &  \textbf{82.8}  & \textbf{87.5}  & \textbf{77.3}  & \textbf{85.7} & 2.17 \\
\bottomrule[1.5pt]
\multicolumn{9}{c}{\textit{Testing 2019 Split}} \\
\midrule[1pt]
MST$^*$~\pub{ICCVW19}~\cite{mst} & & \checkmark &  81.7  &  80.0  &  83.3  &  77.9  &  85.5 & - \\
EMN$^*$~\pub{ICCVW19}~\cite{emn} & & \checkmark  &  81.8  &  \textbf{80.7}  &  84.7  &  77.3  &  84.7 & - \\
\hline
CFBI$^{2\times}$ & & &  81.6  &  80.2  & 84.6 & 77.2  & 84.5 & 0.30 \\
CFBI+$^{2\times}$ & & &  \textbf{82.9}  &  80.6  & \textbf{85.2} & \textbf{78.9}  & \textbf{86.8} & \textbf{0.25} \\

\bottomrule[1.5pt]
\end{tabular}
\end{table}

During training, we firstly downsample all the videos to $480p$ resolution, which is the same as the DAVIS default setting. We adopt SGD with a momentum of $0.9$ and apply a bootstrapped cross-entropy loss, which only considers the $15\%$ hardest pixels. In addition, we apply flipping, scaling, and balanced random-crop as data augmentations. The scaling range is from $1.0$ to $1.3$ times, and the cropped window size is $465\times 465$. During the training stage, we freeze the parameters of BN in the backbone. In the testing stage, all the videos are resized to be no more than $1.3\times480p$ resolution, which is consistent with our training stage. For the multi-scale testing, we apply the scales of $\{1.0, 1.15, 1.3, 1.5\}$ and $\{1.5, 1.7, 1.9\}$ on YouTube-VOS, and DAVIS, respectively. 

For YouTube-VOS experiments, we use a learning rate of $0.01$ for $100,000$ steps with a batch size of 8 using 4 Tesla V100 GPUs. The current sequence's length is $N=3$. The training time on YouTube-VOS is about 3 days. For training with only DAVIS, we use a learning rate of $0.006$ for $50,000$ steps with a batch size of 6 videos using $2$ GPUs. The current sequence's length is $N=3$ as well. For training with both DAVIS and YouTube-VOS, we first train CFBI or CFBI+ on YouTube-VOS following the above setting. After that, we fine-tune the model on DAVIS. To avoid over-fitting, we mix DAVIS videos with YouTube-VOS in a ratio of 1:2 during fine-tuning. Besides, we use a learning rate of $0.01$ for $50,000$ steps with a batch size of 8 videos using 4 Tesla V100 GPUs. The current sequence's length is $N=5$, which is slightly better than $N=3$. We use PyTorch~\cite{pytorch} to implement our method.

\section{Experiments}\label{sec:experiment}

Following the previous state-of-the-art method~\cite{spacetime},
we evaluate our method on YouTube-VOS~\cite{youtubevos}, DAVIS 2016~\cite{davis2016} and DAVIS 2017~\cite{davis2017}. For the evaluation on YouTube-VOS, we train our model on YouTube-VOS training split~\cite{youtubevos}. For DAVIS, we train our model on the DAVIS-2017 training split~\cite{davis2017}. Furthermore, we provide DAVIS results using both DAVIS 2017 and YouTube-VOS for training following some latest works~\cite{feelvos,spacetime}.

The evaluation metric is $\mathcal{J}$ score, calculated as the average IoU between the prediction and the ground truth mask, and  $\mathcal{F}$ score, calculated as an average boundary similarity measure between the boundary of the prediction and the ground truth, and their average value ($\mathcal{J}$\&$\mathcal{F}$). We evaluate our results on the official evaluation server or use official tools.

\subsection{Compare with the State-of-the-art Methods}

\begin{table}[t!]

\caption{The quantitative evaluation on DAVIS 2017~\cite{davis2017}. $^{600p}$: using 600p videos instead of 480p during inference. $^{\ddag}$: timing extrapolated from single-object speed assuming linear scaling in the number of objects.}\label{tab:davis2017}
\begin{center}
\setlength{\tabcolsep}{5pt}
\begin{tabular}{l c c c c c c}
\toprule[1.5pt]
 Methods & F & S  & Avg & $\mathcal{J}$ & $\mathcal{F}$ & t/s \\
\midrule[1pt]
\multicolumn{7}{c}{\textit{Validation Split}} \\
\midrule[1pt]
OSMN~\pub{CVPR18}~\cite{osmn} &   &   &  54.8  & 52.5  & 57.1 & 0.28$^\ddag$ \\
VideoMatch~\pub{ECCV18}~\cite{videomatch} &   &  & 62.4  & 56.5 & 68.2 & 0.35 \\
OnAVOS~\pub{BMVC17}~\cite{onavos} & \checkmark  &   &  63.6  & 61.0  & 66.1 & 26 \\
RGMP~\pub{CVPR18}~\cite{rgmp} &   & \checkmark   & 66.7 & 64.8   & 68.6 & 0.28$^\ddag$ \\
A-GAME~\pub{CVPR19}~\cite{agame} (\textbf{Y}) &   &   & 70.0  &  67.2  & 72.7 & \textbf{0.14}$^\ddag$ \\
FEELVOS~\pub{CVPR19}~\cite{feelvos} (\textbf{Y}) &   &   &  71.5  &  69.1  & 74.0 & 0.51 \\
PReMVOS~\pub{ACCV18}~\cite{premvos} & \checkmark  &   &  77.8  &  73.9  & 81.7 & 37.6 \\
LWLVOS~\pub{ECCV20}~\cite{LWLVOS} (\textbf{Y}) &   &   &  81.6  & 79.1  & 84.1 & ~0.4$^\ddag$ \\
STMVOS~\pub{ICCV19}~\cite{spacetime} (\textbf{Y}) &   & \checkmark  &  81.8  & 79.2  & 84.3 & 0.32$^\ddag$ \\

EGMN~\pub{ECCV20}~\cite{EGMN} (\textbf{Y}) &   & \checkmark  &  82.8  & \textbf{80.2}  & 85.2 & 0.4$^\ddag$ \\
KMNVOS~\pub{ECCV20}~\cite{KMN} (\textbf{Y}) &   & \checkmark  &  82.8  & 80.0  & 85.6 & 0.24$^\ddag$ \\

\hline
CFBI  &   &   &  74.9  & 72.1  & 77.7 & 0.17 \\
CFBI (\textbf{Y}) &   &   &  81.9  & 79.3  & 84.5 & 0.17 \\
CFBI+ (\textbf{Y}) &   &   &  \textbf{82.9}  & 80.1  & \textbf{85.7} & 0.18 \\
CFBI$^{MS}$ (\textbf{Y}) &   &   &  84.2  & 81.6  & 86.8 & 5.94 \\
CFBI+$^{MS}$ (\textbf{Y}) &   &   &  \textbf{84.5}  & \textbf{81.7}  & \textbf{87.3} & 3.94 \\

\bottomrule[1.5pt]
\multicolumn{7}{c}{\textit{Testing Split}} \\
\midrule[1pt]
OSMN~\pub{CVPR18}~\cite{osmn} &   &   &  41.3  & 37.7  & 44.9 & 0.42$^\ddag$ \\
OnAVOS~\pub{BMVC17}~\cite{onavos} & \checkmark  &  &  56.5  & 53.4  & 59.6 & 39 \\
RGMP~\pub{CVPR18}~\cite{rgmp} &   &  \checkmark & 52.9 & 51.3   & 54.4 & 0.42$^\ddag$ \\
FEELVOS~\pub{CVPR19}~\cite{feelvos} (\textbf{Y}) &   &  &  57.8  &  55.2  & 60.5 & 0.54 \\
PReMVOS~\pub{ACCV18}~\cite{premvos}  & \checkmark  &   &  71.6  &  67.5  & 75.7 & 41.3 \\
STMVOS$^{600p}$~\pub{ICCV19}~\cite{spacetime} (\textbf{Y}) &   & \checkmark  &  72.2  & 69.3  & 75.2 & - \\
KMNVOS$^{600p}$~\pub{ECCV20}~\cite{KMN} (\textbf{Y}) &   & \checkmark  &  77.2  & 74.1  & 80.3 & - \\
\hline
CFBI (\textbf{Y})&   &  &  75.0  & 71.4  & 78.7 & \textbf{0.19} \\
CFBI+ (\textbf{Y})&   &  &  75.6  & 71.6  & 79.6 & \textbf{0.19} \\
CFBI$^{600p}$ (\textbf{Y})&   &  &  76.6  & 73.0  & 80.1 & 0.35 \\
CFBI+$^{600p}$ (\textbf{Y})&   &  &  \textbf{78.0}  & \textbf{74.4}  & \textbf{81.6} & 0.29 \\
\bottomrule[1.5pt]
\end{tabular}
\end{center}

\end{table}

{\noindent \textbf{YouTube-VOS}~\cite{youtubevos} is the latest large-scale dataset for multi-object video segmentation. Compared to the popular DAVIS benchmark, which consists of $120$ videos, YouTube-VOS is about 37 times larger. In detail, YouTube-VOS contains 3471 videos in training split (65 categories), 507 videos in validation split (additional 26 unseen categories), and 541 videos in testing split (additional 29 unseen categories). Due to the existence of unseen object categories, the YouTube-VOS validation split is much suitable for measuring the generalization ability of VOS methods. }

{As shown in Table~\ref{tab:youtubevos}, we compare our method with the latest VOS methods on both Validation 2018 and Testing 2019 splits. Without using any bells and whistles, like fine-tuning at test time~\cite{osvos,onavos} or pre-training on larger augmented simulated data~\cite{rgmp,spacetime,EGMN,KMN}, our CFBI+ achieves an average score of ${82.0\%}$, which significantly outperforms all other methods in every evaluation metric.
We can improve the performance of CFBI+ to ${\textbf{82.8}\%}$ (CFBI+$^{2\times}$) by using a stronger training schedule with a double batch size and learning rate. Especially, CFBI+'s multi-object inference speed is much faster than BoLT ($1.36s$).}

{Benefit from the multi-scale matching, our CFBI+ is robuster ($82.0\%$ \vs $81.4\%$) and more efficient ($0.25s$ \vs $0.29s$) than CFBI.
Especially, CFBI+ uses only half of the training batch size to exceed CFBI$^{2\times}$. Besides, the $\mathbf{82.8}\%$ result is significantly higher ($1.4\%$) than KMNVOS, which follows STMVOS to use extensive simulated data for training. Without simulated data, the performance of STMVOS will drop a lot from $79.4\%$ to $68.2\%$. }
{Moreover, we can further boost the performance of CFBI+ to $\mathbf{83.3\%}$ by applying a multi-scale and flip strategy during the evaluation.}

\zongxin{We also compare our method with two of the best results on Testing 2019 split, \ie, \textit{Rank 1} (EMN~\cite{emn}) and \textit{Rank 2} (MST~\cite{mst}) results in the 2nd Large-scale Video Object Segmentation Challenge. Without using model ensemble, simulated data or testing-stage augmentation, our CFBI+ ($\mathbf{82.9\%}$) significantly outperforms the \textit{Rank 1} result ($81.8\%$) while maintaining an efficient multi-object speed of $4$ FPS. Notably, the improvement of CFBI+ mainly comes from the unseen categories ($\mathbf{78.9}\%\mathcal{J}/\mathbf{86.8}\%\mathcal{F}$ \vs $77.3\%\mathcal{J}/84.7\%\mathcal{F}$) instead of seen. Such a strong result further demonstrates CFBI+'s generalization ability and effectiveness.}

\begin{table}[t!]
\centering
\caption{The quantitative evaluation on the DAVIS-2016 validation set~\cite{davis2016}. (\textbf{Y}) denotes additionally using YouTube-VOS for training.}\label{tab:davis2016}
\setlength{\tabcolsep}{6.5pt}
\begin{tabular}{l c c c c c c}
\toprule[1.5pt]
Methods  & F & S  & Avg & $\mathcal{J}$ & $\mathcal{F}$ & t/s \\
\midrule[1pt]
OSMN~\pub{CVPR18}~\cite{osmn} &  &   &  - & 74.0  &   & 0.14 \\
PML~\pub{CVPR18}~\cite{pml} &  &   &  77.4 & 75.5  & 79.3  & 0.28 \\
VideoMatch~\pub{ECCV18}~\cite{videomatch} &  &  &   80.9  & 81.0  & 80.8  & 0.32 \\
RGMP~\pub{CVPR18}~\cite{rgmp} &  &   & 68.8  & 68.6  & 68.9  & 0.14 \\
RGMP~\pub{CVPR18}~\cite{rgmp} &  & \checkmark  & 81.8 &  81.5 & 82.0  & 0.14 \\
A-GAME~\pub{CVPR19}~\cite{agame} (\textbf{Y}) &  &   &  82.1 & 82.2  & 82.0  & \textbf{0.07} \\
FEELVOS~\pub{CVPR19}~\cite{feelvos} (\textbf{Y}) &  &   & 81.7  &  81.1 &  82.2 & 0.45 \\
OnAVOS~\pub{BMVC17}~\cite{onavos}{} & \checkmark &   & 85.0  & 85.7  & 84.2  & 13 \\
PReMVOS~\pub{ACCV18}~\cite{premvos} & \checkmark &   & 86.8  & 84.9  & 88.6  & 32.8 \\
STMVOS~\pub{ICCV19}~\cite{spacetime} (\textbf{Y}) &  & \checkmark  &  89.3 & 88.7  & 89.9  & 0.16 \\
KMNVOS~\pub{ECCV20}~\cite{KMN} (\textbf{Y}) &  & \checkmark  &  \textbf{90.5} & \textbf{89.5}  & \textbf{91.5}  & 0.12 \\
\hline
CFBI  &   &   & 86.1  & 85.3 &  86.9 & 0.16 \\
CFBI (\textbf{Y})  &   &   & 89.4  & 88.3 & 90.5  & 0.16 \\
CFBI+ (\textbf{Y})  &   &   & 89.9  & 88.7 & 91.1  & 0.17 \\
\bottomrule[1.5pt]
\end{tabular}
\end{table}

{\noindent \textbf{DAVIS 2017}~\cite{davis2017} is a multi-object extension of DAVIS 2016. The validation split of DAVIS 2017 consists of 59 objects in 30 videos. And the training split contains 60 videos. Compared to YouTube-VOS, DAVIS is much smaller and easy to be over-fitted.}

{As shown in Table~\ref{tab:davis2017}, CFBI+ exceeds KMNVOS and EGMN ($\mathbf{82.9\%}$ \vs $82.8\%$) without using simulated data. Moreover, CFBI+ achieves a faster multi-object inference speed ($\mathbf{0.18}s$) than KMNVOS ($0.24\%$). 
Different from KMNVOS and EGMN, the backbone features of CFBI+ or CFBI is shared for all the objects in each frame, which leads to a more efficient multi-object inference.
The augmentation in evaluation can further boost CFBI+ to a higher score of $\mathbf{84.5\%}$.}

{We also evaluate our method on the DAVIS-2017 testing split, which is much more challenging than the validation split. On the testing split, we outperform KMNVOS ($77.2\%$) by $\textbf{0.8\%}$ under the setting proposed by STMVOS (\ie, evaluating on $600p$ resolution). When evaluating on the default $480p$ resolution of DAVIS, CFBI+ or CFBI is much better than the STMVOS using $600p$ resolution ($75.6\%$ or $75.0\%$ \vs $72.2\%$). These strong results further prove the generalization ability of CFBI+ and CFBI.}

{The speed of CFBI+ is only comparable with CFBI when evaluating using a $480p$ resolution on DAVIS. The reason for this is that 
convolutional layers have a larger proportion of calculations when evaluating on a small resolution. 
And CFBI+ has more convolutional layers than CFBI because of introducing FPN. On larger resolution (\eg, $600p$), CFBI+ is faster than CFBI ($0.29s$ \vs $0.35s$).}

{\noindent \textbf{DAVIS 2016}~\cite{davis2016} contains 20 videos annotated with high-quality masks each for a single target object. We compare our CFBI method with state-of-the-art methods in Table~\ref{tab:davis2016}. On the DAVIS-2016 validation split, our CFBI+ trained with the additional YouTube-VOS training split achieves an average score of $\textbf{89.9}\%$, which is slightly worse than KMNVOS $90.5\%$, a method using simulated data as mentioned before.
Since the amount of data in DAVIS is tiny, using additional simulation data can help alleviate over-fitting.
Compare to a much fair baseline (\ie, FEELVOS) whose setting is closer to ours, the proposed CFBI+ not only achieves a much better accuracy ($\mathbf{89.9\%}$ \vs $81.7\%$) but also maintains a much faster inference speed ($0.17s$ \vs $0.45s$). }

\subsection{Ablation Study}

\begin{table}[t!]
\centering
\caption{Ablation of background embedding on the DAVIS-2017 validation split. P and I denote the pixel-level matching and instance-level attention, respectively. $^*$: removing the foreground and background bias.}\label{tab:ablation_a}
\setlength{\tabcolsep}{8pt}
\begin{tabular}{l c c c c}
    \toprule[1.5pt]
         P & I & Avg & $\mathcal{J}$ & $\mathcal{F}$ \\
    \midrule[1pt]
        \checkmark  & \checkmark & 74.9 & 72.1 & 77.7 \\
    \hline
        \checkmark$^*$  & \checkmark & 72.8 & 69.5 & 76.1 \\
        \checkmark  &  & 73.0 & 69.9 & 76.0 \\
          & \checkmark & 72.3 & 69.1 & 75.4 \\
          &  & 70.9 & 68.2 & 73.6 \\
    \bottomrule[1.5pt]
\end{tabular}
\end{table}

\begin{table}[t!]
\centering
\caption{Ablation of atrous matching. We evaluate the speed and performance of CFBI on the YouTube-VOS validation split using different atrous matching factors ($l$). $l=1$ is equivalent to original matching.}\label{tab:ablation_am}
\setlength{\tabcolsep}{6.5pt}
\begin{tabular}{l c c c c}
    \toprule[1.5pt]
         $l$ & 1 & 2 & 3 & 4 \\
    \midrule[1pt]
    \multicolumn{5}{c}{\textit{Global Matching}} \\
    \midrule[1pt]
        Avg & 81.4 & 81.3 & 80.7 & 79.9\\

        t/s & 0.29 & 0.15 & 0.13 & 0.12 \\
    \midrule[1pt]
    \multicolumn{5}{c}{\textit{Multi-local Matching}} \\
    \midrule[1pt]
        Avg & 81.4 & 80.8 & 80.1 & 79.5 \\

        t/s & 0.29 & 0.26 & 0.25 & 0.25 \\
    \bottomrule[1.5pt]
\end{tabular}
\end{table}

We analyze the ablation effect of each component proposed in CFBI on the DAVIS-2017 validation split. Following FEELVOS, we use only the DAVIS-2017 training split as training data for these experiments. 

\noindent \textbf{Background Embedding.} As shown in Table~\ref{tab:ablation_a}, we first analyze the influence of removing the background embedding while keeping the foreground only. Without any background mechanisms, the result of our method heavily drops from $74.9\%$ to $70.9\%$.
This result shows that it is significant to embed both foreground and background features collaboratively. Besides, the missing of background information in the pixel-level matching or the instance-level attention will decrease the result to $73.0\%$ or $72.3\%$ separately. 
Thus, compared to instance-level attention, the pixel-level matching performance is more sensitive to the effect of background embedding. A possible reason for this phenomenon is that the possibility of existing some background pixels similar to the foreground is higher than some background instances. Finally, we remove the foreground and background bias, $b_F$ and $b_B$, from the distance metric, and the result drops to $72.8\%$, which further shows that the distance between foreground pixels and the distance between background pixels should be separately considered.

\begin{table}[t!]
\centering
\caption{Ablation of multi-scale matching. We evaluate the speed and performance of our methods on the YouTube-VOS validation split. S: the stride of feature maps. $G$: applying foreground-background global matching. $L$: applying foreground-background multi-local matching. $l$: atrous factor. 
}\label{tab:ablation_msm}
\setlength{\tabcolsep}{5.5pt}
\begin{tabular}{l c c c c c c}
    \toprule[1.5pt]
        Name & S=4 & S=8 & S=16  & Avg & t/s \\
    \midrule[1pt]
        CFBI-S4 & $G$ $L^{l=2}$ & & & 81.6 & 0.65 \\
        CFBI-S4-G2 & $G^{l=2}$ $L^{l=2}$ & & & 81.6 & 0.27 \\
        CFBI-S8 &  & $G$ $L$ & & 80.9 & 0.13 \\
        CFBI-S16 & & & $G$ $L$ & 78.3 & 0.11 \\
        CFBI & $G$ & $L$ & & 81.4 & 0.29 \\
        CFBI-G2 & $G^{l=2}$ & $L$ & & 81.3 & 0.15 \\
        CFBI+ & $G^{l=2}$ $L^{l=2}$ & $G$ $L$ & $G$ $L$ & 82.0 & 0.25 \\
    \bottomrule[1.5pt]
\end{tabular}
\end{table}

\begin{table}[t!]

\centering
\caption{Ablation of other components on the DAVIS-2017 validation split.}\label{tab:ablation_b}
\setlength{\tabcolsep}{8pt}
\begin{tabular}{l c c c c}
    \toprule[1.5pt]
          & Ablation & Avg & $\mathcal{J}$ & $\mathcal{F}$ \\
    \midrule[1pt]
        0  & Ours (CFBI) & 74.9 & 72.1 & 77.7 \\
    \hline
        1  & w/o multi-local windows & 73.8 & 70.8 & 76.8 \\
        2  & w/o sequential training & 73.3 & 70.8 & 75.7 \\
        3  & w/o collaborative ensembler & 73.3 & 70.5 & 76.1 \\
        4  & w/o balanced random-crop & 72.8 & 69.8 & 75.8 \\
        5  & w/o instance-level attention & 72.7 & 69.8 & 75.5 \\
    \hline
        6  & baseline (reproduced FEELVOS) & 68.3 & 65.6 & 70.9 \\
    \bottomrule[1.5pt]
\end{tabular}
\end{table}

\begin{figure*}[t!]
    \centering
    \includegraphics[width=0.9\linewidth]{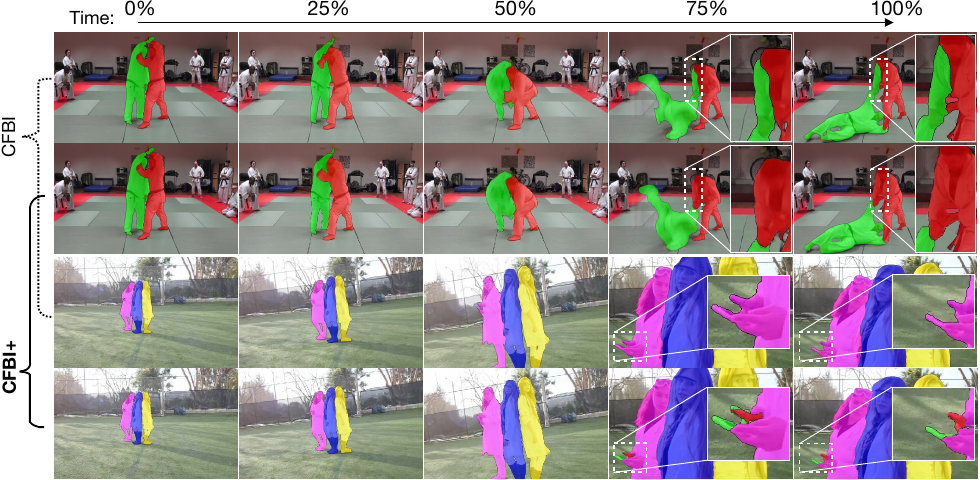}

    \caption{Qualitative comparison between CFBI and CFBI+ on the DAVIS-2017 validation split. In the first video, CFBI fails to segment one hand of the right person (the white box), while CFBI+ generates an accurate boundary between two similar persons. In the second video, CFBI entirely loses two tiny objects (cellphones). In contrast, CFBI+ successfully predicts their masks.}
    \label{fig:comparison}

\end{figure*}

\noindent \textbf{Atrous Matching.} As shown in Table~\ref{tab:ablation_am}, the performance of CFBI will decrease, and the speed will increase as the atrous factor ($l$) increases. Compared to the original matching, 2-atrous matching will significantly accelerate inference speed, but the performance will only slightly decrease. 
The speed will no longer be fast improved by further increasing $l$, and the performance will be heavily decreased. Compared 2-atrous multi-local matching, 2-atrous global matching has a nearly identical performance as the original global matching ($81.3\%$\vs$81.4\%$) but greatly accelerates the speed by $93\%$. 
In short, Atrous Matching can significantly improve the efficiency of matching processes, especially for global matching.

\noindent \textbf{Multi-scale Matching.} 
Table~\ref{tab:ablation_msm} shows the ablation study of multi-scale matching.
In CFBI experiments, the channel dimension of pixel-wise features is $100$. In CFBI+ experiments, the channel dimensions are $32$, $64$, and $128$ for $S=4$, $S=8$, and $S=16$, respectively.
We first evaluate the difference between different matching scales.
As shown, doing matching in larger scales leads to better performance but takes much more inference time. CFBI-S4 is much powerful than CFBI-S16 ($81.6\%$ \vs $78.3\%$). However, CFBI-S16 is about $5$ times faster than CFBI-S4. For better efficiency, CFBI brings the multi-local matching of CFBI-S4 from $S=4$ to $S=8$, which improves $124\%$ speed and loses only $0.2\%$ performance. If we want to do matching on larger scales, Atrous global matching ($l=2$) is critic in saving computational resources (CFBI-G2 $0.15s$ \vs CFBI $0.29s$, CFBI-S4-G2 $0.27s$ \vs CFBI-S4 $0.65s$) while losing little performance (CFBI-G2 $81.3\%$ \vs CFBI $81.4\%$, CFBI-S4-G2 $81.6\%$ \vs CFBI-S4 $81.6\%$). 
Finally, by combining all the matching processes on three scales and progressively increasing their channel dimensions, CFBI+ achieves better performance ($82.0\%$) while proposed atrous matching helps guarantee an efficient speed ($0.25s$).

\begin{figure*}[t!]
    \centering
    \includegraphics[width=0.9\linewidth]{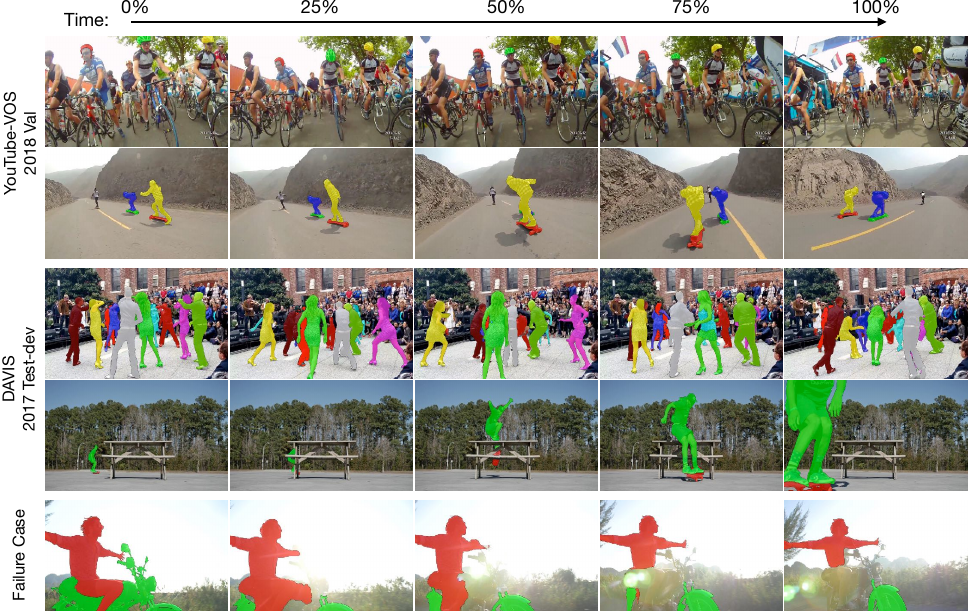}
    \caption{Qualitative results of CFBI+ on the DAVIS-2017 testing split and YouTube-VOS 2018 validation split. These videos cover many of the most challenging VOS cases, including similar objects, small objects, occlusion, and blur. In the first four videos, CFBI+ performs well on similar objects, small objects, and occlusion. Especially for the third video, there are $10$ similar targets (dancers) in total and many similar people in the background. Besides, all the targets continuously occlude each other. However, CFBI+ does not collapse under such a complicated case and correctly track every dancer. In the last video, we show one of the worst cases on the DAVIS-2017 testing split, where CFBI+ fails to segment all the motorbike parts. The blur caused by such a strong halo makes it difficult for CFBI+ to distinguish the motorbike's appearance.}
    \label{fig:quality}
\end{figure*}

\noindent \textbf{Qualitative Comparison.} To further compare CFBI with CFBI+, we visualize some representative comparison results on the DAVIS-2017 validation split in Fig.~\ref{fig:comparison}. Benefit from the local matching on larger scales, CFBI+ can generate a more accurate boundary between similar targets. Moreover, CFBI+ is capable of predicting some tiny objects, which are difficult for CFBI. In addition, we show more results of CFBI+ under some of the hardest cases on the DAVIS-2017 testing split and YouTube-VOS in Fig.~\ref{fig:quality}. CFBI generalize well on most of these cases, including similar objects, small objects, and occlusion.

\noindent \textbf{Other Components.} The ablation study of other proposed components is shown in Table~\ref{tab:ablation_b}. Line 0 ($74.9\%$) is the result of proposed CFBI, and Line 6 ($68.3\%$) is our baseline method reproduced by us. Under the same setting, our CFBI significantly outperforms the baseline.

In line 1, we use only one local neighborhood window to conduct the local matching following the setting of FEELVOS, which degrades the result from $74.9\%$ to $73.8\%$. It demonstrates that our multi-local matching module is more robust and effective than the single-local matching module of FEELVOS. Notably, the computational complexity of multi-local matching dominantly depends on the biggest local window size because we use the intermediate results of the local matching of the biggest window to calculate on smaller windows.

In line 2, we replace our sequential training by using ground-truth masks instead of network predictions as the previous mask. By doing this, the performance of CFBI drops from $74.9\%$ to $73.3\%$, which shows the effectiveness of our sequential training under the same setting.

In line 3, we replace our collaborative ensembler with 4 depth-wise separable convolutional layers (and we keep applying instance-level attention before each separable convolutional layer). This architecture is the same as the dynamic segmentation head of~\cite{feelvos}. Compared to our collaborative ensembler, the dynamic segmentation head has much smaller receptive fields and performs $1.6\%$ worse.

In line 4, we use normal random-crop instead of our balanced random-crop during the training process. In this situation, the performance drops by $2.1\%$ to $72.8\%$ as well. As expected, our balanced random-crop is successful in relieving the model form biasing to background attributes.

In line 5, we disable the use of instance-level attention as guidance information to the collaborative ensembler, which means we only use pixel-level information to guide the prediction. In this case, the result deteriorates even further to $72.7$, which proves that instance-level information can further help the segmentation with pixel-level information.

In summary, we explain the effectiveness of each proposed component of CFBI and CFBI+. For VOS, it is necessary to embed both foreground and background features. Besides, the model will be more robust by combining pixel-level information and instance-level information. Notably, multi-scale pixel-level matching is much more potent than single-scale, and atrous matching is critical for improving matching efficiency. Apart from this, the proposed balanced random-crop and sequential training are useful but straightforward in improving training performance.

\section{Conclusion} \label{sec:conclusion}

This paper proposes a novel framework for video object segmentation by introducing collaborative foreground-background integration and achieves new state-of-the-art results on three popular benchmarks. Specifically, we impose the feature embedding from the foreground target and its corresponding background to be contrastive. Moreover, we integrate both pixel-level and instance-level embeddings to make our framework robust to various object scales while keeping the network simple and fast. 
In particular, our design of multi-scale matching can further improve VOS's performance, and our atrous matching can greatly improve the efficiency of matching processes. We hope CFBI or CFBI+ will serve as a solid baseline and help ease the future research of VOS and related areas, such as video object tracking and interactive video editing.

\noindent \textbf{Acknowledgements.} This work is partly supported by ARC DP200100938 and ARC DECRA DE190101315.


%



\ifCLASSOPTIONcaptionsoff
  \newpage
\fi



\small{
\bibliographystyle{IEEEtran}
\bibliography{reference}

\begin{thebibliography}{10}
\providecommand{\url}[1]{#1}
\csname url@samestyle\endcsname
\providecommand{\newblock}{\relax}
\providecommand{\bibinfo}[2]{#2}
\providecommand{\BIBentrySTDinterwordspacing}{\spaceskip=0pt\relax}
\providecommand{\BIBentryALTinterwordstretchfactor}{4}
\providecommand{\BIBentryALTinterwordspacing}{\spaceskip=\fontdimen2\font plus
\BIBentryALTinterwordstretchfactor\fontdimen3\font minus
  \fontdimen4\font\relax}
\providecommand{\BIBforeignlanguage}[2]{{%
\expandafter\ifx\csname l@#1\endcsname\relax
\typeout{** WARNING: IEEEtran.bst: No hyphenation pattern has been}%
\typeout{** loaded for the language `#1'. Using the pattern for}%
\typeout{** the default language instead.}%
\else
\language=\csname l@#1\endcsname
\fi
#2}}
\providecommand{\BIBdecl}{\relax}
\BIBdecl

\bibitem{ngan2011video}
K.~N. Ngan and H.~Li, \emph{Video segmentation and its applications}.\hskip 1em
  plus 0.5em minus 0.4em\relax Springer Science \& Business Media, 2011.

\bibitem{zhang2016instance}
Z.~Zhang, S.~Fidler, and R.~Urtasun, ``Instance-level segmentation for
  autonomous driving with deep densely connected mrfs,'' in \emph{CVPR}, 2016,
  pp. 669--677.

\bibitem{vis}
L.~Yang, Y.~Fan, and N.~Xu, ``Video instance segmentation,'' in \emph{ICCV},
  2019, pp. 5188--5197.

\bibitem{Feng_2019_ICCV}
Q.~Feng, Z.~Yang, P.~Li, Y.~Wei, and Y.~Yang, ``Dual embedding learning for
  video instance segmentation,'' in \emph{ICCV Workshops}, 2019.

\bibitem{oh2019fast}
S.~W. Oh, J.-Y. Lee, N.~Xu, and S.~J. Kim, ``Fast user-guided video object
  segmentation by interaction-and-propagation networks,'' in \emph{CVPR}, 2019,
  pp. 5247--5256.

\bibitem{miao2020memory}
J.~Miao, Y.~Wei, and Y.~Yang, ``Memory aggregation networks for efficient
  interactive video object segmentation,'' in \emph{CVPR}, 2020.

\bibitem{liangmemory}
C.~Liang, Z.~Yang, J.~Miao, Y.~Wei, and Y.~Yang, ``Memory aggregated cfbi+ for
  interactive video object segmentation,'' in \emph{CVPR Workshops}, 2020.

\bibitem{osvos}
S.~Caelles, K.-K. Maninis, J.~Pont-Tuset, L.~Leal-Taix{\'e}, D.~Cremers, and
  L.~Van~Gool, ``One-shot video object segmentation,'' in \emph{CVPR}, 2017,
  pp. 221--230.

\bibitem{onavos}
P.~Voigtlaender and B.~Leibe, ``Online adaptation of convolutional neural
  networks for video object segmentation,'' in \emph{BMVC}, 2017.

\bibitem{premvos}
J.~Luiten, P.~Voigtlaender, and B.~Leibe, ``Premvos: Proposal-generation,
  refinement and merging for video object segmentation,'' in \emph{ACCV}, 2018,
  pp. 565--580.

\bibitem{osmn}
L.~Yang, Y.~Wang, X.~Xiong, J.~Yang, and A.~K. Katsaggelos, ``Efficient video
  object segmentation via network modulation,'' in \emph{CVPR}, 2018, pp.
  6499--6507.

\bibitem{feelvos}
P.~Voigtlaender, Y.~Chai, F.~Schroff, H.~Adam, B.~Leibe, and L.-C. Chen,
  ``Feelvos: Fast end-to-end embedding learning for video object
  segmentation,'' in \emph{CVPR}, 2019, pp. 9481--9490.

\bibitem{spacetime}
S.~W. Oh, J.-Y. Lee, N.~Xu, and S.~J. Kim, ``Video object segmentation using
  space-time memory networks,'' in \emph{ICCV}, 2019.

\bibitem{voc}
M.~Everingham, L.~Van~Gool, C.~K. Williams, J.~Winn, and A.~Zisserman, ``The
  pascal visual object classes (voc) challenge,'' \emph{IJCV}, vol.~88, no.~2,
  pp. 303--338, 2010.

\bibitem{coco}
T.-Y. Lin, M.~Maire, S.~Belongie, J.~Hays, P.~Perona, D.~Ramanan,
  P.~Doll{\'a}r, and C.~L. Zitnick, ``Microsoft coco: Common objects in
  context,'' in \emph{ECCV}.\hskip 1em plus 0.5em minus 0.4em\relax Springer,
  2014, pp. 740--755.

\bibitem{cheng2014global}
M.-M. Cheng, N.~J. Mitra, X.~Huang, P.~H. Torr, and S.-M. Hu, ``Global contrast
  based salient region detection,'' \emph{TPAMI}, vol.~37, no.~3, pp. 569--582,
  2014.

\bibitem{shi2015hierarchical}
J.~Shi, Q.~Yan, L.~Xu, and J.~Jia, ``Hierarchical image saliency detection on
  extended cssd,'' \emph{TPAMI}, vol.~38, no.~4, pp. 717--729, 2015.

\bibitem{semantic}
B.~Hariharan, P.~Arbel{\'a}ez, L.~Bourdev, S.~Maji, and J.~Malik, ``Semantic
  contours from inverse detectors,'' in \emph{ICCV}.\hskip 1em plus 0.5em minus
  0.4em\relax IEEE, 2011, pp. 991--998.

\bibitem{davis2016}
F.~Perazzi, J.~Pont-Tuset, B.~McWilliams, L.~Van~Gool, M.~Gross, and
  A.~Sorkine-Hornung, ``A benchmark dataset and evaluation methodology for
  video object segmentation,'' in \emph{CVPR}, 2016, pp. 724--732.

\bibitem{davis2017}
J.~Pont-Tuset, F.~Perazzi, S.~Caelles, P.~Arbel{\'a}ez, A.~Sorkine-Hornung, and
  L.~Van~Gool, ``The 2017 davis challenge on video object segmentation,''
  \emph{arXiv preprint arXiv:1704.00675}, 2017.

\bibitem{youtubevos}
N.~Xu, L.~Yang, Y.~Fan, D.~Yue, Y.~Liang, J.~Yang, and T.~Huang, ``Youtube-vos:
  A large-scale video object segmentation benchmark,'' \emph{arXiv preprint
  arXiv:1809.03327}, 2018.

\bibitem{cfbi}
Z.~Yang, Y.~Wei, and Y.~Yang, ``Collaborative video object segmentation by
  foreground-background integration,'' in \emph{ECCV}, 2020.

\bibitem{xiao2018monet}
H.~Xiao, J.~Feng, G.~Lin, Y.~Liu, and M.~Zhang, ``Monet: Deep motion
  exploitation for video object segmentation,'' in \emph{CVPR}, 2018, pp.
  1140--1148.

\bibitem{masktrack}
F.~Perazzi, A.~Khoreva, R.~Benenson, B.~Schiele, and A.~Sorkine-Hornung,
  ``Learning video object segmentation from static images,'' in \emph{CVPR},
  2017, pp. 2663--2672.

\bibitem{flownet}
A.~Dosovitskiy, P.~Fischer, E.~Ilg, P.~Hausser, C.~Hazirbas, V.~Golkov, P.~Van
  Der~Smagt, D.~Cremers, and T.~Brox, ``Flownet: Learning optical flow with
  convolutional networks,'' in \emph{ICCV}, 2015, pp. 2758--2766.

\bibitem{favos}
J.~Cheng, Y.-H. Tsai, W.-C. Hung, S.~Wang, and M.-H. Yang, ``Fast and accurate
  online video object segmentation via tracking parts,'' in \emph{CVPR}, 2018,
  pp. 7415--7424.

\bibitem{pml}
Y.~Chen, J.~Pont-Tuset, A.~Montes, and L.~Van~Gool, ``Blazingly fast video
  object segmentation with pixel-wise metric learning,'' in \emph{CVPR}, 2018,
  pp. 1189--1198.

\bibitem{videomatch}
Y.-T. Hu, J.-B. Huang, and A.~G. Schwing, ``Videomatch: Matching based video
  object segmentation,'' in \emph{ECCV}, 2018, pp. 54--70.

\bibitem{rgmp}
S.~Wug~Oh, J.-Y. Lee, K.~Sunkavalli, and S.~Joo~Kim, ``Fast video object
  segmentation by reference-guided mask propagation,'' in \emph{CVPR}, 2018,
  pp. 7376--7385.

\bibitem{EGMN}
X.~Lu, W.~Wang, M.~Danelljan, T.~Zhou, J.~Shen, and L.~Van~Gool, ``Video object
  segmentation with episodic graph memory networks,'' in \emph{ECCV}, 2020.

\bibitem{KMN}
H.~Seong, J.~Hyun, and E.~Kim, ``Kernelized memory network for video object
  segmentation,'' in \emph{ECCV}, 2020.

\bibitem{LWLVOS}
G.~Bhat, F.~J. Lawin, M.~Danelljan, A.~Robinson, M.~Felsberg, L.~Van~Gool, and
  R.~Timofte, ``Learning what to learn for video object segmentation,'' in
  \emph{ECCV}, 2020.

\bibitem{attention_conv1}
J.~Gehring, M.~Auli, D.~Grangier, D.~Yarats, and Y.~N. Dauphin, ``Convolutional
  sequence to sequence learning,'' in \emph{ICML}, 2017.

\bibitem{attention_conv2}
Y.~N. Dauphin, A.~Fan, M.~Auli, and D.~Grangier, ``Language modeling with gated
  convolutional networks,'' in \emph{ICML}, 2017.

\bibitem{senet}
J.~Hu, L.~Shen, and G.~Sun, ``Squeeze-and-excitation networks,'' in
  \emph{CVPR}, 2018.

\bibitem{deeplabv3p}
L.-C. Chen, Y.~Zhu, G.~Papandreou, F.~Schroff, and H.~Adam, ``Encoder-decoder
  with atrous separable convolution for semantic image segmentation,'' in
  \emph{ECCV}, 2018, pp. 801--818.

\bibitem{fpn}
T.-Y. Lin, P.~Doll{\'a}r, R.~Girshick, K.~He, B.~Hariharan, and S.~Belongie,
  ``Feature pyramid networks for object detection,'' in \emph{CVPR}, 2017, pp.
  2117--2125.

\bibitem{torralba2003contextual}
A.~Torralba, ``Contextual priming for object detection,'' \emph{IJCV}, vol.~53,
  no.~2, pp. 169--191, 2003.

\bibitem{resnet}
K.~He, X.~Zhang, S.~Ren, and J.~Sun, ``Deep residual learning for image
  recognition,'' in \emph{CVPR}, 2016.

\bibitem{deeplab}
L.-C. Chen, G.~Papandreou, I.~Kokkinos, K.~Murphy, and A.~L. Yuille, ``Deeplab:
  Semantic image segmentation with deep convolutional nets, atrous convolution,
  and fully connected crfs,'' \emph{TPAMI}, vol.~40, no.~4, pp. 834--848, 2017.

\bibitem{pspnet}
H.~Zhao, J.~Shi, X.~Qi, X.~Wang, and J.~Jia, ``Pyramid scene parsing network,''
  in \emph{CVPR}, 2017, pp. 2881--2890.

\bibitem{holschneider1990real}
M.~Holschneider, R.~Kronland-Martinet, J.~Morlet, and P.~Tchamitchian, ``A
  real-time algorithm for signal analysis with the help of the wavelet
  transform,'' in \emph{Wavelets}.\hskip 1em plus 0.5em minus 0.4em\relax
  Springer, 1990, pp. 286--297.

\bibitem{dilated_conv}
F.~Yu and V.~Koltun, ``Multi-scale context aggregation by dilated
  convolutions,'' \emph{arXiv preprint arXiv:1511.07122}, 2015.

\bibitem{xception}
F.~Chollet, ``Xception: Deep learning with depthwise separable convolutions,''
  in \emph{CVPR}, 2017, pp. 1251--1258.

\bibitem{bn}
S.~Ioffe and C.~Szegedy, ``Batch normalization: Accelerating deep network
  training by reducing internal covariate shift,'' in \emph{ICML}, 2015.

\bibitem{deng2009imagenet}
J.~Deng, W.~Dong, R.~Socher, L.-J. Li, K.~Li, and L.~Fei-Fei, ``Imagenet: A
  large-scale hierarchical image database,'' in \emph{CVPR}.\hskip 1em plus
  0.5em minus 0.4em\relax Ieee, 2009, pp. 248--255.

\bibitem{gn}
Y.~Wu and K.~He, ``Group normalization,'' in \emph{ECCV}, 2018, pp. 3--19.

\bibitem{gct}
Z.~Yang, L.~Zhu, Y.~Wu, and Y.~Yang, ``Gated channel transformation for visual
  recognition,'' 2020.

\bibitem{agame}
J.~Johnander, M.~Danelljan, E.~Brissman, F.~S. Khan, and M.~Felsberg, ``A
  generative appearance model for end-to-end video object segmentation,'' in
  \emph{CVPR}, 2019, pp. 8953--8962.

\bibitem{boltvos}
P.~Voigtlaender, J.~Luiten, and B.~Leibe, ``Boltvos: Box-level tracking for
  video object segmentation,'' \emph{arXiv preprint arXiv:1904.04552}, 2019.

\bibitem{mst}
Q.~Zhou, Z.~Huang, L.~Huang, Y.~Gong, H.~Shen, W.~Liu, and X.~Wang,
  ``Motion-guided spatial time attention for video object segmentation,'' in
  \emph{ICCV Workshops}, 2019.

\bibitem{emn}
Z.~Zhou, L.~Ren, P.~Xiong, Y.~Ji, P.~Wang, H.~Fan, and S.~Liu, ``Enhanced
  memory network for video segmentation,'' in \emph{ICCV Workshops}, 2019.

\bibitem{pytorch}
A.~Paszke, S.~Gross, S.~Chintala, G.~Chanan, E.~Yang, Z.~DeVito, Z.~Lin,
  A.~Desmaison, L.~Antiga, and A.~Lerer, ``Automatic differentiation in
  pytorch,'' 2017.

\end{thebibliography}
}
%


%
\vspace{-12mm}
\begin{IEEEbiography}[{\includegraphics[width=0.9in,clip,keepaspectratio]{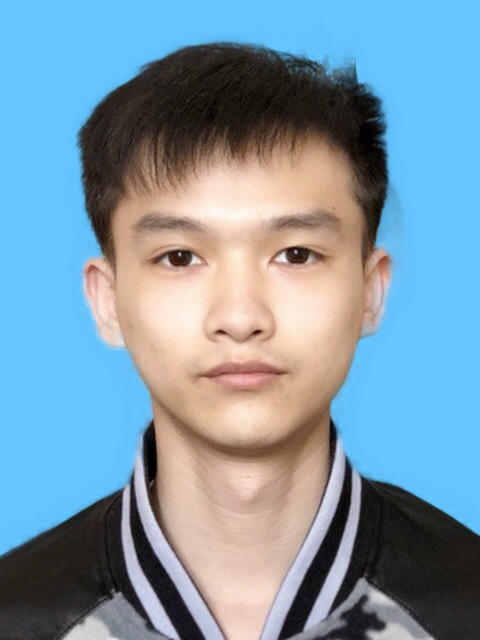}}]{Zongxin Yang} is currently a Ph.D. student with the University of Technology Sydney. He received his Bachelor degree from University of Science and Technology of China, Hefei, China, in 2018. His current research interest is computer vision, including image recognition and video object segmentation.
\end{IEEEbiography}
\vspace{-16mm}
\begin{IEEEbiography}[{\includegraphics[width=0.9in,clip,keepaspectratio]{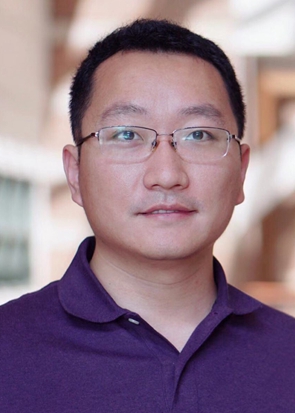}}]{Yunchao Wei} is currently an Assistant Professor with the University of Technology Sydney. He received his Ph.D. degree from Beijing Jiaotong University, Beijing, China, in 2016. He was a Postdoctoral Researcher at Beckman Institute, UIUC, from 2017 to 2019. He is ARC Discovery Early Career Researcher Award Fellow from 2019 to 2021. His current research interests include computer vision and machine learning.
\end{IEEEbiography}
\vspace{-16mm}
\begin{IEEEbiography}[{\includegraphics[width=0.9in,clip,keepaspectratio]{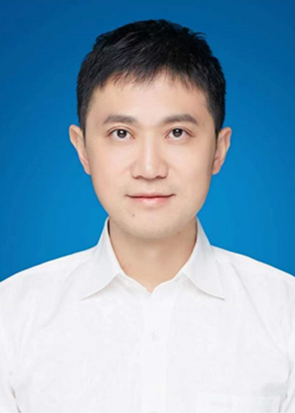}}]{Yi Yang} is currently a Professor with the University of Technology Sydney. He received his Ph.D degree from Zhejiang University, Hangzhou, China, in 2010. He was a post-doctoral researcher in the School of Computer Science, Carnegie Mellon University. His current research interests include machine learning and its applications to multimedia content analysis and computer vision, such as multimedia retrieval and video content understanding.
\end{IEEEbiography}




\end{document}


%
\title{Supplementary Materials for ``Self-Correction for Human Parsing"}
%
%
%
%

\author{Peike~Li,
        Yunqiu~Xu,
        Yunchao~Wei,~\IEEEmembership{Member,~IEEE}
        and~Yi~Yang\IEEEauthorrefmark{1},~\IEEEmembership{Senior~Member,~IEEE}
}

%
%

\markboth{IEEE TRANSACTIONS ON PATTERN ANALYSIS AND MACHINE INTELLIGENCE, 2020.}%
{Shell \MakeLowercase{\textit{et al.}}: Bare Demo of IEEEtran.cls for Computer Society Journals}
%


\IEEEtitleabstractindextext{%

}

\maketitle
\IEEEdisplaynontitleabstractindextext
%
\IEEEpeerreviewmaketitle

\input{Section/appendix}

\ifCLASSOPTIONcaptionsoff
  \newpage
\fi

\clearpage
\bibliographystyle{IEEEtran}
\bibliography{./reference.bib}
